# Histogram Transporter: Learning Rotation-Equivariant Orientation Histograms for High-Precision Robotic Kitting

Jiadong Zhou, Yadan Zeng, Huixu Dong, and I-Ming Chen, *Fellow, IEEE*

*Abstract*—Robotic kitting is a critical task in industrial automation that requires the precise arrangement of objects into kits to support downstream production processes. However, when handling complex kitting tasks that involve fine-grained orientation alignment, existing approaches often suffer from limited accuracy and computational efficiency. To address these challenges, we propose Histogram Transporter, a novel kitting framework that learns high-precision pick-and-place actions from scratch using only a few demonstrations. First, our method extracts rotation-equivariant orientation histograms (EOHs) from visual observations using an efficient Fourier-based discretization strategy. These EOHs serve a dual purpose: improving picking efficiency by directly modeling action success probabilities over high-resolution orientations and enhancing placing accuracy by serving as local, discriminative feature descriptors for object-to-placement matching. Second, we introduce a subgroup alignment strategy in the place model that compresses the full spectrum of EOHs into a compact orientation representation, enabling efficient feature matching while preserving accuracy. Finally, we examine the proposed framework on the simulated Hand-Tool Kitting Dataset (HTKD), where it outperforms competitive baselines in both success rates and computational efficiency. Further experiments on five Raven-10 tasks exhibits the remarkable adaptability of our approach, with real-robot trials confirming its applicability for real-world deployment.

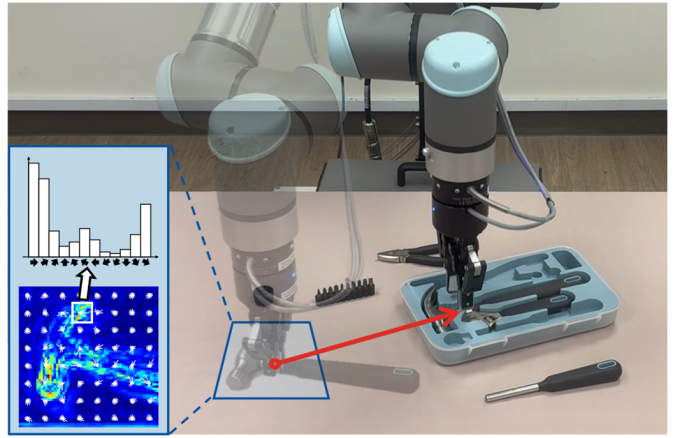

Fig. 1. Histogram Transporter learns to grasp tools and precisely place them into corresponding kit cavities with tight clearances by leveraging pixel-wise EOH encodings derived from visual inputs. The bottom left shows an example EOH map for the hammer, where EOHs are depicted as multi-directional arrows at a stride of eight, with pixel colors indicating the maximal EOH value at each corresponding pixel.

## I. INTRODUCTION

PRECISE robotic manipulation is essential for industrial automation, enabling various production activities such as material feeding, kitting, bin picking, and assembly [1]-[3]. In this work, we focus on robotic kitting, the process of selecting related components and arranging them into a container as a single product unit [4]. Effective kitting operations streamline downstream production while reducing storage and shipping costs, but achieving these benefits requires high-precision robotic operation, especially when dealing with irregularly shaped, elongated parts or cavities with tight clearances. As shown in Fig. 1 for the hand-tool kitting task, even minor positional or angular errors can prevent tools from fitting into corresponding conformal cavities.

Robotic kitting has attracted significant research interest in recent years. Traditional methods estimate object poses within a scene to guide kitting actions [5], [6]. These approaches rely on strong prior knowledge of objects, such as high-quality 3D models or considerable training data, along with customized engineering for object-specific manipulation planning. This is particularly problematic for high-mix, low-volume kitting lines, where new products are frequently introduced [4]. In contrast, alternative approaches utilize end-to-end policies that map visual observations directly to robot actions using imitation learning [7], [8]. While these methods simplify system pipelines and adapt well to diverse tasks, including kitting, they often require substantial robot-environment interactions for data collection and struggle to be generalized to new configurations, such as varying object poses.

Recent progress in policy learning has exploited underlying symmetries in robotic manipulation tasks to improve training efficiency. A notable example is the Equivariant Transporter [9], [10], a behavior-cloning method that incorporates bi-

The project is supported by A*STAR under "RIE2025 IAF-PP Advanced ROS2-native Platform Technologies for Cross Sectorial Robotics Adoption (M21K1a0104)" programme.

J. Zhou, Y. Zeng and I-M. Chen are with the School of Mechanical and Aerospace Engineering, Nanyang Technological University, Singapore (email: {jiadong001, yadan001}@e.ntu.edu.sg, michen@ntu.edu.sg).

H. Dong is with the Grasp Laboratory, Mechanical Engineering Department, Zhejiang University, Hangzhou 310027, China (e-mail: huixudong@zju.edu.cn).

equivariance in planar pick-and-place tasks, ensuring that transformations of objects and their target placement poses lead to corresponding transformations in the pick-and-place action distribution. These properties are incorporated as inductive biases in model architectures via equivariant neural networks [11]. Despite showcasing exceptional sample efficiency and generalizability, these methods have primarily been applied to simple manipulation scenarios, such as rearranging cubic blocks, where low orientation precision suffices. Although angular resolution is often treated as a task-specific parameter, the effectiveness of these methods in orientation-sensitive tasks, such as kitting hand tools, remains uncertain. Furthermore, scaling to larger rotation spaces significantly increases model parameters, resulting in greater computational complexity and memory consumption.

To this end, we aim to develop an effective kitting framework capable of learning orientation-sensitive tasks in a data-efficient manner. To achieve this, we propose Histogram Transporter, a novel end-to-end imitation learning framework for high-precision robotic kitting. Building on Equivariant Transporter, our framework inherits its sample efficiency and generalizability while enhancing orientation scalability and discriminability by introducing rotation-equivariant orientation histograms (EOHs) for visual encoding. These EOHs are end-to-end trainable feature descriptors that preserve equivariant structures under image rotations and enrich orientation information across multiple discrete angles. To efficiently generate fine-grained EOHs for high-precision tasks, we employ a Fourier-based method that models feature distributions over orientations with band-limited functions, enabling continuous sampling at any angle. Within our framework, EOHs serve two key functions: in the pick module, a single EOH vector directly represents action success distributions across orientations, while in the place module, dense, pixel-wise EOHs act as local discriminative feature descriptors for establishing visual correspondence between the target object and its intended placement region. To enhance scalability, the place model incorporates a subgroup alignment strategy to map the full orientation information of EOHs into a more compact representation, optimizing computational efficiency without compromising accuracy.

Finally, we conduct extensive experiments on both simulated and real-robot platforms to validate the accuracy and computational efficiency of Histogram Transporter. Using hand-tool kitting as a case study, our method outperforms Equivariant Transporter on the Hand-Tool Kitting Dataset (HTKD) [12], achieving a 96% success rate on unseen tool poses with only 10 demonstrations, and 92% on unseen tool instances with 100 demonstrations, while requiring $5.1 \times$ less memory and delivering $1.4 \times$ faster inference. Beyond kitting, the proposed method also demonstrates state-of-the-art performance on Raven-10 tasks, showcasing remarkable data efficiency and adaptability. To further evaluate its practical applicability, we deploy our method on a UR10e robot, successfully imitating three real-world tasks using only 10 human demonstrations.

Our main contributions are summarized as follows.
- We propose rotation-equivariant orientation histograms (EOHs) as deep features with strong orientation discriminability and introduce an efficient Fourier-based discretization method for fine-grained EOH generation.
- We present Histogram Transporter, an imitation learning approach for high-precision kitting that uses EOHs to model pick success probabilities over discrete orientations and act as local feature descriptors for precise placement matching.
- We incorporate a subgroup alignment strategy in the place model to generate compact EOH encodings, improving computational efficiency in deep feature template matching.
- We validate our method through extensive experiments in both simulation and real-world settings, demonstrating its accuracy, scalability, and adaptability.

The rest of this paper is organized as follows. Section II reviews related work. Section III presents the necessary preliminaries on equivariant learning and formulates the problem. Section IV introduces the proposed methodology, detailing EOHs and their application in Histogram Transporter. Section V reports evaluation results on HTKD and real-world kitting experiments, while additional pick-and-place experiments are discussed in Section VI. Finally, Section VII concludes the paper.

II. RELATED WORK

*A. Vision-based Robotic Kitting*

A classical approach to vision-based robotic kitting relies on accurate pose estimation of object parts to guide subsequent robot action planning [13]-[16]. For instance, Domae *et al.* [5] utilized an edge-based pose estimation method based on 3D CAD models to kit industrial parts into desired poses within a tray. Similarly, Shi and Koonjul [6] combined real-time pose estimation of known 3D objects with grasp planning for dexterous robotic hands to execute kitting tasks. While effective, these methods often depend on high-quality CAD models, which may not always be available in real-world scenarios. Alternately, other approaches focus on kitting previously unseen objects without relying on 3D models. Devgon *et al.* [17] trained a convolutional neural network to estimate relative rotations between objects and their matching cavities, while Li *et al.* [18] proposed an action-snapping algorithm to refine rough kitting poses provided by human operators. However, these methods often require large simulation datasets for training, posing sim-to-real challenges. In contrast to these methods based on inferred object poses, our approach is object-agnostic, directly predicting end-effector poses from only a few demonstrations.

Recent studies have explored end-to-end policy learning for pick-and-place tasks, offering a convenient and versatile learning paradigm for kitting and other manipulation tasks [19]. Zakka *et al.* [7] studied generalizable kit assembly by learning dense shape descriptors that establish geometric correspondence between object surfaces and their target cavities. Zeng *et al.* [8] introduced Transporter Networks, addressing placement tasks through deep feature template matching between target objects and their intended placement regions. However, a common drawback of these approaches is the mismatch between pixel-level positioning and relatively coarse orientation, often limited to small rotation spaces containing only 20 [7], 32 [20] or 36 [8] discrete angles. Some studies have utilized coarse-to-fine-grain schemes to achieve high-precision actions in large action spaces [3], [21]. For example, Gualtieri and Platt [22] employed a hierarchical spatial attention mechanism that progressively zooms into smaller parts of the scene to produce high-dimensional, precise robot actions. Sóti et al. [23] proposed an iterative inference method to incrementally reduce rotation errors in Transporter Networks. In contrast, our method infers precise kitting policies over high-resolution translation and rotation spaces in a single pass, offering a streamlined and computationally efficient solution.

*B. Learning Equivariance in Manipulation*

Fully Convolutional Networks (FCNs) have demonstrated remarkable learning efficiency in various vision-based manipulation tasks due to their translationally equivariant structures [24], [25]. Recent advancements have extended these capabilities to rotational equivariance through Equivariant Convolutional Networks. This concept is first introduced with G-CNNs [26] and later generalized to broader rotation groups via Steerable CNNs [27]. Weiler and Cesa [11] developed E(2)-Steerable CNNs as a comprehensive framework for the Euclidean group E(2) and its subgroups, while Cesa et al. [28] proposed E(3)-steerable kernels using a band-limited harmonic basis. These architectures have proven effective across diverse vision domains, such as keypoint description [29] and oriented object detection [30].

Equivariant models have emerged as promising approaches for robotic manipulation, significant improving sample efficiency and generalizability. Wang et al. [20], [31] explored SO(2)-equivariant reinforcement learning for multi-step pick-and-place tasks, while Zhu et al. [32] incorporated SE(2)-equivariance into online grasp planning. Of particular relevance to our work, the Equivariant Transporter [9], [10] leveraged bi-equivariance in planar pick-and-place tasks, and the recently proposed Fourier Transporter [33] extended this framework to 3D manipulation by modeling action distributions with truncated Fourier representations. Building on similar principles, Zhou et al. [12] tackled high-precision kitting tasks through fine-grained and efficient rotation discretization. Unlike these methods, which determine place actions by matching two rotation-invariant encodings (target object and intended placement region), our approach employs EOHs —rotation-equivariant encodings with high sensitivity to rotations. Additionally, Ryu et al. [34] introduced high-degree equivariant point features to enhance generalizability in manipulation learning, similar to our method. However, while their approach relies heavily on pose sampling to optimize the energy-based model, our method directly predicts the action distribution over the entire space, enabling more efficient learning for planar kitting tasks.

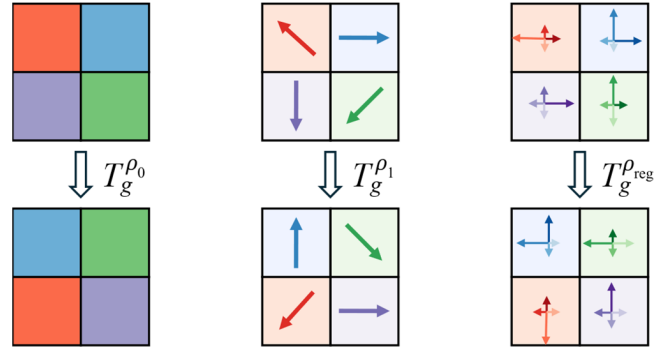

Fig. 2. Illustration of transformations $T_g^\rho$ applied to $2 \times 2$ feature fields associated with representations $\rho_0, \rho_1$, and $\rho_{\text{reg}}$ for a $\pi/2$ CCW rotation ($g = \pi/2$). Each $T_g^\rho$ combines a spatial rotation (determined by $\rho_1$, acting on pixel coordinates) and a group-wise rotation (determined by $\rho$). Left: $\rho_0$ leaves the 1-channel feature map invariant. Middle: $\rho_1$ rotates the 2-channel vector at each pixel. Right: $\rho_{\text{reg}}$ permutes the order of the 4-channel feature map, where the arrows correspond to group channels arranged in the sequence $[0, \pi/2, \pi, 3\pi/2]$).

## III. PRELIMINARIES & PROBLEM FORMULATION

*A. Planar Rotation Group and Representation*

We are primarily concerned with planar rotations characterized by the group SO(2) and its cyclic subgroup $C_N <$ SO(2). SO(2) represents the continuous group of all planar rotations, defined as SO(2) = $\{\text{Rot}_\theta : 0 \leq \theta < 2\pi\}$. In contrast, the subgroup $C_N = \{\text{Rot}_\theta : \theta \in \{2\pi i/N \mid 0 \leq i < N\}\}$ contains N discrete rotations by multiples of $2\pi/N$. While SO(2) encompasses infinite elements, $C_N$ consists of exactly $|C_N| = N$ elements.

A representation $\rho : G \to GL(\mathbb{R}^D)$ of a group G maps each group element $g \in G$ to an invertible $D \times D$ matrix $\rho(g) \in GL(\mathbb{R}^D)$. We consider four specific representations that describe how vectors $v \in \mathbb{R}^D$ transform under planar rotations. The size of the vectors, $|v| = D$, is same as the dimension of $\rho$, denoted $|\rho| = D$.

1) The *trivial representation* $\rho_0$ acts on a scalar $v \in \mathbb{R}$ by leaving it unchanged: $\rho_0(g)v = v$ for all $g \in G$.
2) The *standard representation* $\rho_1$ rotates a vector $v \in \mathbb{R}^2$ according to a standard rotation matrix associated with $g$, as detailed in Eq. (1) with $j = 1$. $\rho_0$ and $\rho_1$ are valid for both SO(2) and $C_N$.
3) The *regular representation* $\rho_{\text{reg}}$ of $C_N$ cyclically permutes elements of $v \in \mathbb{R}^N$: $\rho_{\text{reg}}(g)(v_0, v_1, \ldots, v_{N-1}) = (v_{N-i}, \ldots, v_{N-1}, v_0, v_1, \ldots, v_{N-1-i})$, where $g$ is the $i$-th element of $C_N$.
4) The *irreducible representations* (irreps) of SO(2) include the trivial representation $\rho_0$ and a series of $2 \times 2$ rotation matrices:

$$\rho_j(\text{Rot}_\theta) = \begin{bmatrix} \cos(j\theta) & -\sin(j\theta) \\ \sin(j\theta) & \cos(j\theta) \end{bmatrix}, \quad (1)$$

where $j \in \mathbb{N}^+$ denotes the frequency of the irrep. Any SO(2) representation $\rho$ can be block-diagonalized as a direct sum of the irreps: $\rho(g) = Q^T(\bigoplus_{j \in J} \rho_j)Q$, where $Q$ is an orthogonal change-of-basis matrix, and $J$ specifies the involved irreps. The irreps of $C_N$ are subsets of those of SO(2), restricted to $j \in$

$[0, (N-1)/2]$.

*B. Feature Fields and Their Transformation*

In equivariant CNNs [11], [28], 2D feature maps are interpreted as feature vector fields that transform according to a specified representation $\rho$ of a group G. Unlike standard CNNs, where a K-channel feature map is represented as a tensor $f \in \mathbb{R}^{K \times H \times W}$, equivariant CNNs introduce an additional group dimension to encode transformations in the feature space. This modification results in a feature map $f \in \mathbb{R}^{K \times D \times H \times W}$, with $D = |\rho|$ indicating the dimension of the group representation. Conceptually, this can be viewed as a stack of K feature fields, each represented as $f_k : \mathbb{R}^{H \times W} \to \mathbb{R}^D$, mapping every pixel $x \in \mathbb{R}^{H \times W}$ to a vector $f_k(x) \in \mathbb{R}^D$ for $0 \leq k < K$. Here, K is referred to as the feature dimension, D as the group dimension, and $H \times W$ as the spatial dimension.

A group element $g \in G$ acts on feature vector fields through the transformation $T_g^\rho$:

$$[T_g^\rho(f)](x) = \rho(g) \cdot f(\rho_1(g)^{-1}x), \quad (2)$$

where $x \in \mathbb{R}^2$ represents continuous spatial dimensions to facilitate the transformation analysis. In Eq. (2), $\rho_1$ rotates the pixel coordinates in the spatial dimension, while $\rho \in \{\rho_0, \rho_1, \rho_{reg}, \rho_{irrep}\}$ transforms each pixel's feature vector $f_k(x) \in \mathbb{R}^D$ in the group dimension. Illustrative examples of transformations under different $\rho$-type are provided in Fig. 2.

*C. Equivariance and Rotation-Equivariant Convolution*

A function $F : X \to Y$ is considered equivariant with respect to a group G if it commutes with the actions of G on $X$ and $Y$:

$$F(T_g^{\rho_X}(f)) = T_g^{\rho_Y}(F(f)) \quad \forall (f, g) \in (X, G), \quad (3)$$

where $T_g^{\rho_X}$ and $T_g^{\rho_Y}$ denote the group actions on $X$ and $Y$, respectively. While standard convolutional operations are inherently translation-equivariant, rotational equivariance can be achieved using G-steerable kernels $\kappa(x) \in \mathbb{R}^{|\rho_Y| \times |\rho_X|}$ that satisfy the steerability constraint [27].

A rotation-equivariant network can be constructed by sequentially stacking rotation-equivariant convolutional layers. Let $F = \{L_i | i \in \{1, 2, 3, \ldots, l\}\}$ denote a set of $l$ equivariant layers under a group G, the equivariance of the entire network can be expressed as:

$$\left[\prod_{i=1}^{l} L_i\right](T_g^{\rho_{in}}(I)) = T_g^{\rho_{out}}\left(\left[\prod_{i=1}^{l} L_i\right](I)\right), \quad (4)$$

where $I$ represents an input image, and $T_g^{\rho_{in}}$ and $T_g^{\rho_{out}}$ denote the transformations applied to the input and output of the network under the group action by $g \in G$. Notably, this equivariance constraint also holds for residual networks with skip connections.

*D. Problem Definition*

In this paper, we formulate robotic kitting as a high-precision planar pick-and-place problem within the framework of behavior cloning. Building on the techniques in [9], we incorporate a significantly higher angular resolution in action parameterization to achieve precise orientation alignment. Formally, the kitting problem is defined as follows: given a visual observation $o_t$ as input, the objective is to determine high-resolution actions $a_t = (a_{pick}, a_{place})$ for a parallel-jaw gripper, enabling accurate object placement in its target position and orientation within the kit. The observation $o_t \in \mathbb{R}^{d \times H \times W}$ is an orthographic heightmap of the workspace, reconstructed from RGB-D images captured by overhead cameras. The actions $a_{pick}$ and $a_{place}$ specify the gripper poses for executing the respective pick and place operations. These actions are determined by optimizing a policy $p(a_t|o_t)$ learnt from expert demonstrations. Specifically, this policy is factorized into $p(a_{pick}|o_t)$ and $p(a_{place}|o_t, a_{pick})$, which are modeled using separate neural networks in our approach.

To practically implement the policy networks, we adopt a discretized action space, $\widehat{SE}(2) = \mathbb{Z}^2 \times C_N$, for both $a_{pick}$ and $a_{place}$. Each action is parameterized as a tuple $(u, v, \theta) \in \widehat{SE}(2)$, where the positional component $(u, v) \in \{1, \ldots, H\} \times \{1, \ldots, W\} \subset \mathbb{Z}^2$ corresponds to pixel coordinates in $o_t$. The angular component $\theta \in \{2\pi i/N\}_{i=0}^{N-1}$ represents one of N orientations uniformly spaced across $2\pi$, each corresponding to an element $g \in C_N$. The orientation number N is a critical parameter that defines the angular resolution of actions, directly influencing the precision of rotational predictions. In particular, $\theta_{pick}$ is defined relative to the world frame, indicating absolute pick angles, while $\theta_{place}$ is the orientation change between the pick and place poses, denoting the angular adjustment required for an accurate placement.

IV. METHODOLOGY

This section presents our proposed methodology, which centers on EOHs and their application in Histogram Transporter for high-precision robotic kitting. Subsection IV-A introduces EOHs, covering their definition, transformation properties, Fourier-based generation, and equivariance characteristics. The following subsections detail Histogram Transporter, which leverages EOHs in two key aspects of the kitting process: the pick angle policy is modeled using a single EOH vector (Subsection IV-B), while the place policy is derived through deep feature template matching with EOH maps (Subsection IV-C). Subsection IV-C also presents the subgroup alignment strategy, and additional implementation details are provided in Subsection IV-D.

*A. Rotation-Equivariant Orientation Histograms*

**1) Definition and properties of EOHs.** Given an input image $I$, our approach encodes it into an EOH map $O$, which preserves an equivariant structure while enriching orientation information. The map $O \in \mathbb{R}^{|C_N| \times H \times W}$ is a tensor where each pixel $x \in \mathbb{R}^{H \times W}$ is assigned a discrete, normalized EOH vector $O(x) \in \mathbb{R}^{|C_N|}$. This vector spans $|C_N| = N$ orientation bins and encodes local orientation information at each pixel using a discrete orientation probability distribution. The interpretation of this probabilistic orientation information is task-dependent. For example, in Subsection IV-B, EOHs are used to model pick rotation distributions. Fig. 3 visualizes the EOH maps of a hammer, showing a dense grid of multi-directional arrows (spaced every eight pixels for clarity). Each arrow corresponds to a bin value of an EOH vector, with its orientation indicating the associated bin angle in a circular layout.

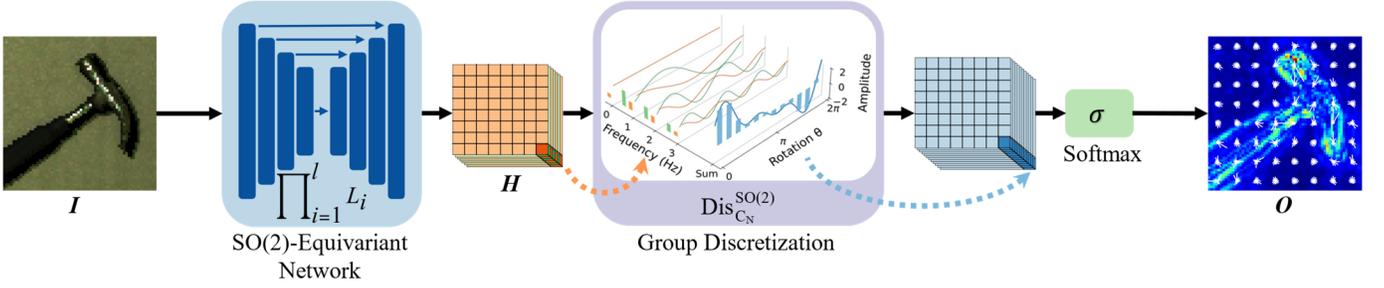

Fig. 3. Fourier-based EOH generation process. An input observation $I$ passes through an SO(2)-equivariant network to produce pixel-wise coefficient vectors $H$ in the Fourier domain. These vectors are then transformed into the discrete orientation domain via group discretization. Finally, a softmax operation normalizes the discrete orientation vectors, generating the dense pixel-wise EOH map $O$.

In the context of equivariance learning, an EOH map can be interpreted as a feature field $O : \mathbb{R}^{H \times W} \to \mathbb{R}^{|C_N|}$ associated with the regular representation $\rho_{\text{reg}}$ of $C_N$. By substituting $O$ into Eq. (2), the transformed map under a rotation $g \in C_N$ is represented as:

$$\left[T_g^{\rho_{\text{reg}}}(O)\right](x) = \rho_{\text{reg}}(g) \cdot O(\rho_1(g)^{-1}x), \quad (5)$$

where the transformation $T_g^{\rho_{\text{reg}}}$ combines the spatial rotation $\rho_1$ in pixel coordinates with the cyclic permutation $\rho_{\text{reg}}$ in the orientation bins. This transformation mechanism, illustrated in Fig. 4 (right column), shows how the EOH map rotates such that the densely distributed multi-directional arrows remain rigidly attached. This property ensures a coherent transformation across both spatial and group dimensions.

*2) Fourier-based EOH generation.* Fig. 3 provides an overview of the Fourier-based EOH generation process. We utilize an SO(2)-equivariant network to extract Fourier-based orientation representations. The network layers are defined with respect to the representation:

$$\rho_{\text{irrep}} = \bigoplus_{j \in [0, j_c]} \rho_j, \quad (6)$$

where $\rho_j$ denotes the irreducible representations of SO(2) up to frequency $j_c \in \mathbb{N}^+$. The dimension of $\rho_{\text{irrep}}$ is given by $|\rho_{\text{irrep}}| = 1 + 2j_c$. The first layer $L_1$ maps the $\rho_0$ field of the input image to $\rho_{\text{irrep}}$ fields, while subsequent layers operate within $\rho_{\text{irrep}}$. Given an input image $I \in \mathbb{R}^{d \times H \times W}$, the $l$-layer equivariant network produces an output field $H$:

$$H = \left[\prod_{i=1}^{l} L_i\right](I), \quad (7)$$

where $H \in \mathbb{R}^{|\rho_{\text{irrep}}| \times H \times W}$ assigns each pixel $x \in \mathbb{R}^{H \times W}$ a coefficient vector $H(x) \in \mathbb{R}^{|\rho_{\text{irrep}}|}$. Intermediate layers $L_{i \neq l}$ generate field stacks $H_{i \neq l} \in \mathbb{R}^{K_i \times |\rho_{\text{irrep}}| \times H \times W}$, while the final layer $L_l$ collapses the feature dimension $K_l$ to produce $H$. Each vector $H(x)$ contains Fourier coefficients corresponding to sine and cosine basis functions of SO(2) at frequencies specified by $\rho_{\text{irrep}}$. These coefficients reconstruct a continuous SO(2)-signal at each pixel via the Inverse Fourier Transform $\mathcal{F}^{-1}$:

$$\mathcal{F}^{-1}[H(x)](g) = a_0 + \sum_{j=1}^{j_c}[a_j\cos(jg) + b_j\sin(jg)], \quad (8)$$

where $g$ is an arbitrary angle in SO(2); $a_j$ and $b_j$ are the Fourier coefficients in $H(x)$ for frequency $j$. This formulation enables compact and precise orientation representation at each pixel.

To obtain a discrete EOH representation, we sample $H(x)$

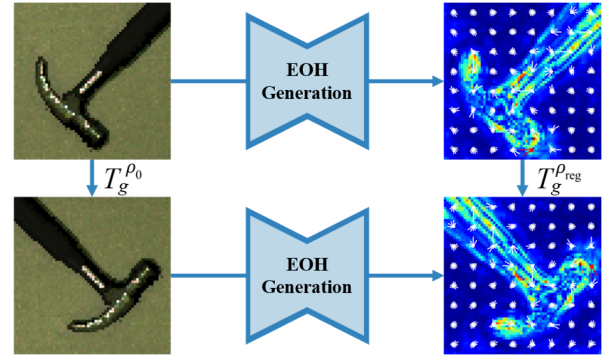

Fig. 4. Rotation equivariance in the EOH generation process.

across predefined discrete orientations using Eq. (8). This process, termed group discretization, maps Fourier coefficients to a discrete $C_N$-signal:

$$\text{Dis}_{C_N}^{SO(2)}\bigl(H(x)\bigr) = \{\mathcal{F}^{-1}[H(x)](g)\}_{\forall g \in C_N}. \quad (9)$$

Here, $\mathcal{F}^{-1}$ is applied to each orientation $g$ in $C_N$, effectively transforming an SO(2) $\rho_{\text{irrep}}$-vector into a $C_N$ $\rho_{\text{reg}}$-vector at every position $x$ in $H$. The final EOH map $O$ is computed as:

$$O = \sigma\left(\text{Dis}_{C_N}^{SO(2)}(H)\right), \quad (10)$$

where $\sigma$ denotes a pixel-wise softmax function that normalizes the histogram vectors, converting raw histogram data into a probabilistic format. This pixel-wise EOH vector provides a detailed description of the local orientation distribution, forming a fundamental component of our kitting framework.

*3) Efficiency and equivariance of the EOH generation process.* The EOH generation process provides an efficient and flexible way to produce orientation histograms with varying bin numbers. By employing a consistent set of irreducible representations $\rho_{\text{irrep}}$ in the SO(2)-equivariant network, the structure of the network remains independent of the desired orientation precision. Instead, the orientation resolution is controlled by adjusting the sampling rate during the final group discretization step. This discretization process is efficiently implemented via a simple multiplication with a change-of-basis matrix $Q \in \mathbb{R}^{|C_N| \times |\rho_{\text{irrep}}|}$, thus minimizing computational overhead. However, to prevent signal aliasing, the sampling number N must satisfy $N \geq |\rho_{\text{irrep}}|$, in accordance with the Nyquist-Shannon sampling theorem.

Our Fourier-based EOH generation method $F : I \to O$ maintains rotation equivariance and satisfies the following

constraint:
$$F(T_g^{\rho_0}(I)) = T_g^{\rho_{\text{reg}}}(F(I)) \quad \forall g \in C_N, \quad (11)$$
where a rotation $g$ applied to the input image $I$ results in a corresponding rotation of the output EOH map $O = F(I)$. Due to discretization, the rotation action is restricted to the discrete group $C_N$, with the degree of approximation to full SO(2)-equivariance determined by the chosen sampling rate N. Fig. 4 illustrates the rotation-equivariance property of the EOH generation process.

*B. Direct Modelling of Pick Policy*

While an EOH map could theoretically encode the pick policy $p(a_{\text{pick}}|o_t)$ over entire action space $\widehat{SE}(2)$, we adopt the augmented state representation (ASR) [9], [20], which has shown superior performance in planar manipulation tasks. As a result, we focus on modeling the angular component of the policy with an EOH vector for efficient and accurate orientation prediction.

*1) ASR in planar picking.* The ASR method decomposes the original pick policy into two parts: a translational component $p(u, v)$ and an angular component $p(\theta|(u, v))$. These components are separately modeled using two neural networks: $f_p(o_t) \to p(u, v)$ and $f_\theta(o_t, (u, v)) \to p(\theta|(u, v))$.

The pick position model $f_p : \mathbb{R}^{d \times H \times W} \to \mathbb{R}^{1 \times H \times W}$ maps a d-channel input observation $o_t$ to a 1-channel probability map $p(u, v)$, representing the picking success probabilities at pixel coordinates $(u, v)$. Since $f_p$ is independent of pick orientations, the group dimension of the model output is pooled out, making the position model invariant to rotations.

The pick angle model $f_\theta : \mathbb{R}^{d \times H_1 \times W_1} \to \mathbb{R}^{|C_N|}$ takes as input a cropped region of $o_t$ centered at $(u, v)$ and outputs a probability vector $p(\theta|(u, v)) \in \mathbb{R}^{|C_N|}$, representing the picking success probability over N discrete orientations at $(u, v)$. The probability vector is effectively an EOH vector, encoding the success probability for each orientation bin. It can be viewed as a special case of the EOH map $O \in \mathbb{R}^{|C_N| \times H \times W}$ with H = W = 1.

The optimal pick action $a_{\text{pick}}^*$ is determined by maximizing the pick policy $p(a_{\text{pick}}|o_t)$. Specifically, the optimal pick position is identified as $(u^*, v^*) = \arg\max_{(u, v)} p(u, v)$, and the optimal pick angle is subsequently determined as: $\theta^* = \arg\max_\theta p(\theta|(u^*, v^*))$. Fig. 5 illustrates this inference process, with the top part showcasing the evaluation of $f_p$ and the bottom part depicting the evaluation of $f_\theta$.

*2) Equivariance learning in planar picking.* Both $f_p$ and $f_\theta$ are constructed using rotation-equivariant convolutional layers to preserve the equivariant symmetry inherent in planar picking tasks, as proposed in [9]. Specifically, both models utilize $\rho_0$ as the input field type and employ irreducible representations $\rho_{\text{irrep}}$ in their intermediate layers. However, their output field types differ: $f_p$ applies a group pooling operation on the $\rho_{\text{irrep}}$ field to produce a $\rho_0$ field, while $f_\theta$ uses a group discretization operation to yield a $\rho_{\text{reg}}$ field. Notably, the architecture of $f_\theta$ is a practical implementation of the EOH generation method described in Subsection IV-A, facilitating

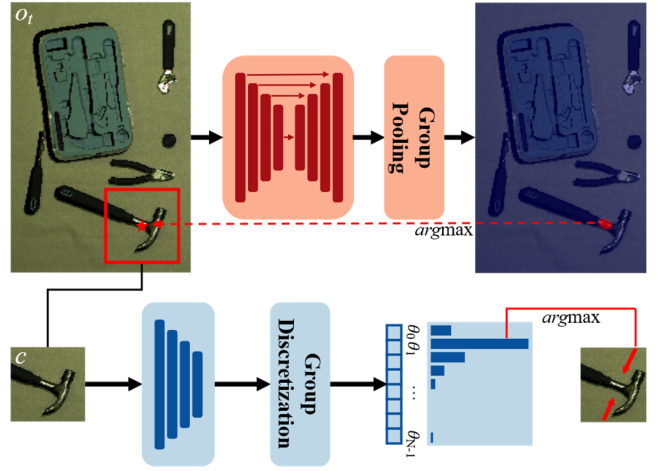

Fig. 5. Overview of the pick module. The pick position model $f_p$: The input observation $o_t$ is processed by an equivariant U-Net followed by a group pooling operation to generate the pick probability map, determining the optimal pick position (top row). The pick angle model $f_\theta$: A crop $c$, centered at the pick position, is input to an equvariant ResNet, followed by a group discretization operation, producing an EOH vector that indicates pick success probabilities over orientations (bottom row).

efficient policy construction with high angular resolutions.

Consequently, the equivariant constraints enforced in our pick models can be formally expressed as follows:
$$f_p\left(T_g^{\rho_0}(o_t)\right) = T_g^{\rho_0}\left(f_p(o_t)\right) \forall g \in SO(2), \quad (12)$$
$$f_\theta\left(T_g^{\rho_0}(o_t), T_g^{\rho_0}(u, v)\right) = \rho_{\text{reg}}(g) f_\theta\left(o_t, (u, v)\right) \forall g \in C_N. \quad (13)$$
The above equations indicate that a rotational transformation $T_g^{\rho_0}$ on the input observation $o_t$ results in corresponding transformations of the pick position and angle distributions.

*3) Gripper symmetry with the quotient group.* The bilateral symmetry of a parallel-jaw gripper introduces additional invariance in planar picking, as the grasp success probability remains unchanged when the grasp pose is rotated by $\pi$ around the forward axis. To incorporate this invariance, the pick angle model $f_\theta$ is adapted to use the quotient group $SO(2)/C_2$, which treats rotations differing by $\pi$ as equivalent. Specifically, the irreducible representations $\rho_{\text{irrep}}$ in Eq. (6) are redefined with respective to $SO(2)/C_2$ by employing basis functions with a period of $\pi$. The group discretization operation is also adjusted to sample N/2 orientations within $[0, \pi)$, aligning with the elements of $C_N/C_2$. This adaptation naturally accounts for the gripper's rotational symmetry, reducing redundancy in the action space and enhancing the model's computational efficiency.

*C. Template Matching and Subgroup Alignment for Place Policy*

Given the pick action $a_{\text{pick}}$, the place model $f_{\text{place}}$ determines the place action $a_{\text{place}}$ to transport the object to its intended kitting pose. Assuming the object remains stationary during picking, $a_{\text{pick}}$ can be geometrically represented by an image crop $c$ centered at the pick position. The model $f_{\text{place}} : (o_t, c) \to p(a_{\text{place}}|o_t, a_{\text{pick}})$ predicts the probability that an object grasped at $a_{\text{pick}}$ in scene $o_t$ should be placed at $a_{\text{place}}$ to complete the kitting task.

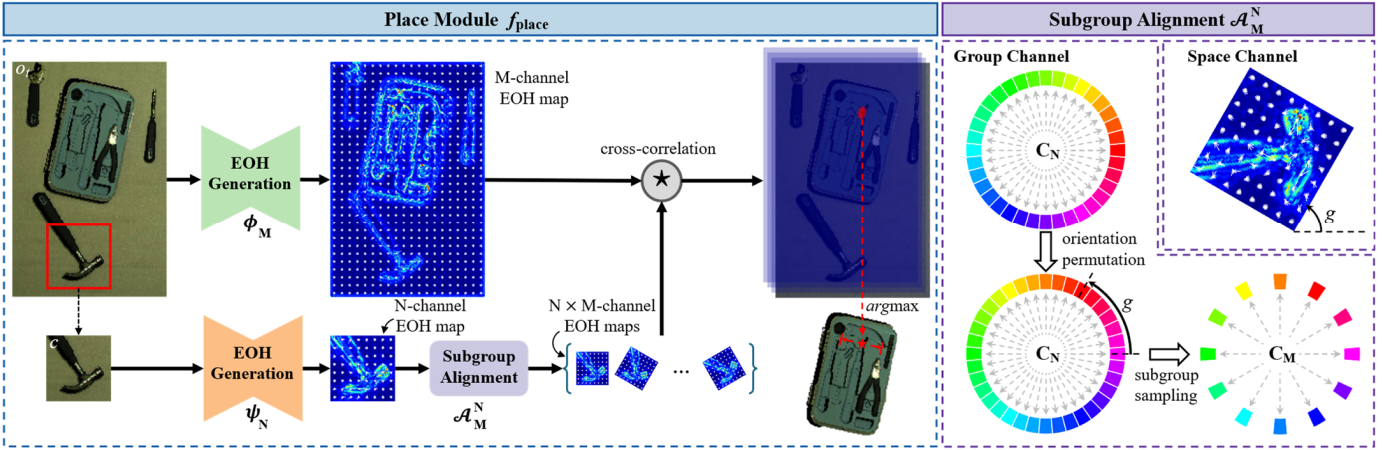

Fig. 6. Overview of the place module. Left: The place model processes the observation $o_t$ and the crop $c$, centered at the pick position, encoding them into M-channel and N-channel EOH maps via the EOH generation networks, $\phi_M$ and $\psi_N$, respectively. A subgroup alignment process is applied to the crop encoding to produce a batch of N rotated EOH maps with compact M group channels. The place action distribution is then computed by cross-correlating the EOH maps of $o_t$ and $c$. Right: The subgroup alignment strategy combines spatial rotation (pixel coordinates) with orientation-wise rotation (group channels). The group channels are cyclically permuted according to the group element $g$, followed by a subgroup sampling operation to achieve a compact representation.

Inspired by [8], [9], our place model $f_{\text{place}}$ employs a template matching approach to establish visual correspondence between the pick action $a_{\text{pick}}$ and the target placement $a_{\text{place}}$. Unlike previous methods, our approach encodes both the scene $o_t$ and the image crop $c$ into EOH maps, enhancing the orientation discriminability of the model. To balance computational efficiency and high orientation precision, we integrate a subgroup alignment strategy into the place model.

*1) EOH encoding.* Two equivariant networks, $\phi_N$ and $\psi_N$, encode the observation $o_t$ and the image crop $c$ into $|C_N|$-channel EOH maps, where N specifies the orientation number in the action space. These networks implement the EOH generation process detailed in Subsection IV-A. The resulting EOH maps, $\phi_N(o_t)$ and $\psi_N(c)$, retain the spatial dimensions of their respective inputs, with each pixel endowed with an EOH vector that serves as a local discriminative feature descriptor.

Unlike Equivariant Transporter [9], which uses group-pooled, rotation-invariant descriptors, our approach retains the group dimension within EOH maps, offering two key advantages. First, it enhances representational capacity by capturing network responses across multiple orientations. While invariant descriptors rely solely on spatial relations among pixels to infer an object's orientation, our descriptors incorporate additional pixel-wise orientation information, enabling improved orientation discrimination and matching. Second, the EOH map systematically encodes orientation variations across both spatial and group dimensions, as defined in Eq. (5). This structured interplay between these dimensions increases sensitivity to orientation changes. As a result, the proposed EOH encodings exhibit strong discriminative power, making them well-suited for orientation-sensitive tasks.

*2) Subgroup alignment.* Although EOH encodings offer superior orientation discriminability, encoding and matching the full $C_N$ orientation spectrum becomes computationally prohibitive as the number of orientation bins N increases. Instead of maintaining a large rotation group with N orientations, a subgroup alignment strategy is introduced to map the feature space to a smaller subgroup $C_M \leq C_N$. This allows the network to represent orientation information compactly, ensuring computational efficiency while still capturing essential rotation features.

To implement this strategy, we subsample EOH maps to transition their group dimension from $C_N$ to its subgroup $C_M$. Given an EOH map $H \in \mathbb{R}^{|C_N| \times H \times W}$, the subsampling operation $S_{C_M}^{C_N} : \mathbb{R}^{|C_N|} \to \mathbb{R}^{|C_M|}$ reduces the EOH vector $H(x) \in \mathbb{R}^{|C_N|}$ at each pixel $x \in \mathbb{R}^{H \times W}$ to a compact form:

$$S_{C_M}^{C_N}(H(x)) = \{[H(x)](g)\}_{\forall g \in C_M}. \tag{14}$$

The resulting M-channel EOH map $S_{C_M}^{C_N}(H) \in \mathbb{R}^{|C_M| \times H \times W}$ remains lightweight while preserving equivariance.

Fig. 6 illustrates how subgroup alignment rotates and shrinks EOH maps to generate more compact templates ($N = 36$ and $M = 12$). Specifically, the EOH map $\psi_N(c)$ undergoes rotational transformations $T_g^{\rho_{\text{reg}}}$ for each $g \in C_N$. After each rotation, the transformed map is subsampled using Eq. (14), aligning it with $C_M$. This process results in a stack of N rotated M-channel EOH maps:

$$\mathcal{A}_M^N(\psi_N(c)) = \{S_{C_M}^{C_N}[T_g^{\rho_{\text{reg}}}(\psi_N(c))]\}_{\forall g \in C_N}. \tag{15}$$

Here, $\mathcal{A}_M^N$ denotes the subgroup alignment strategy acting on the crop's EOH map across all rotations $g \in C_N$.

*3) Place model construction.* Once the subgroup-aligned EOH map stack $\mathcal{A}_M^N(\psi_N(c))$ is generated, the place model $f_{\text{place}}$ is constructed through cross-correlation with $\phi_N(o_t)$:

$$\begin{aligned} f_{\text{place}}(o_t, c) &= \mathcal{A}_M^N(\psi_N(c)) \star S_{C_M}^{C_N}[\phi_N(o_t)] \\ &= \mathcal{A}_M^N(\psi_N(c)) \star \phi_M(o_t), \end{aligned} \tag{16}$$

where $\phi_N(o_t)$ is subsampled accordingly to align its group channels with $C_M$. The subsampling operation $S_M^N$ is seamlessly integrated with the network $\phi_N$ in Eq. (16), resulting in a refined network $\phi_M$ that directly outputs an EOH map associated with $C_M$. The resulting place distribution $p(a_{\text{place}}|o_t, a_{\text{pick}})$ can also be viewed as an EOH map, where each bin value represents the success rates of the corresponding place action. The optimal place action $a^*_{\text{place}}$ is identified by jointly maximizing this distribution across both spatial and orientation dimensions.

Importantly, the choice of $C_M$ does not alter the output dimensions of the place distribution. Instead, using a smaller subgroup $C_M < C_N$ provides an approximation to the full $C_N$ group. Our experiments show that tuning M appropriately can balance accuracy and computational efficiency. Furthermore, to avoid aliasing, M must satisfy the Nyquist-Shannon sampling theorem [28].

*4) Bi-equivariance in planar placing.* Leveraging the template-matching paradigm and the equivariant properties of EOH maps, our place model satisfies the bi-equivariance constraint [9] inherent in planar placing tasks. Specifically, the model $f_{\text{place}}$ exhibits $C_M \times C_M$-equivariance:

$$f_{\text{place}}\left(T^{\rho_0}_{g_1}(c), T^{\rho_0}_{g_2}(o_t)\right) = T^{\rho_{\text{reg}}}_{g_2}\rho_{\text{reg}}(g_1^{-1}) f_{\text{place}}(c, o_t), \quad (17)$$

where $g_1, g_2 \in C_M$, act on the object crop $c$ and the scene observation $o_t$, respectively. This bi-equivariance property ensures that the output place distribution from $f_{\text{place}}$ transforms consistently in response to the independent rotations applied to the object and the scene.

### D. Model Implementation Details

Both the pick and place models are implemented in PyTorch [36] and utilize two rotation-equivariant networks. The equivariant convolution layers in these networks are constructed using the ESCNN library [28].

*1) Pick position model $f_p$.* This model is a 21-layer convolution network that predicts a dense, pixel-wise probability map $p(u, v) \in \mathbb{R}^{H \times W}$, representing the picking success rate at each pixel, based on an RGB-D input $o_t \in \mathbb{R}^{4 \times H \times W}$. The network adopts a U-Net backbone with 9 residual blocks, each containing two equivariant convolution layers and a skip connection. The encoder achieves a stride of 16 using 4 two-stride max-pooling layers, while the decoder restores spatial resolution through 4 bilinear-upsampling layers. All equivariant convolution layers are built on irreducible representations $\rho_{\text{irrep}} = \bigoplus_{j \in [0, j_c=3]} \rho_j$ under SO(2). The first layer maps the input $o_t$, modeled as a trivial field, to $\rho_{\text{irrep}}$ for subsequent processing within the U-Net. The U-Net output is converted back to a trivial field using group average pooling [28], followed by refinement through two $1 \times 1$ standard convolutional layers and SoftMax normalization across the image. To maintain equivariance while introducing nonlinearity, pointwise regular ELU activations [28] are interleaved with the equivariant convolution layers throughout the network.

*2) Pick angle model $f_\theta$.* This model is a 9-layer residual network that maps an image crop $c \in \mathbb{R}^{4 \times H_1 \times W_1}$, centered at the picking location $(u^*, v^*)$, to an orientation distribution $p(\theta|(u^*, v^*)) \in \mathbb{R}^{|C_N|/2}$. The network's first layer transforms the trivial representation of $c$ into irreducible representations $\rho_{\text{irrep}} = \bigoplus_{j \in [0, j_c=6]} \rho_j$ under SO(2). This is followed by three residual blocks, each consisting of a max-pooling layer and two equivariant convolution layers. An additional convolution layer then converts the resulting $\rho_{\text{irrep}}$ vector into quotient representations relative to $SO(2)/C_2$. Finally, a group discretization operation samples the desired orientation values within $C_N/C_2$, producing the discrete orientation probabilities.

*3) Place models $\phi_M$ and $\psi_N$.* These models are 20-layer equivariant networks with a similar architecture to $f_p$, with key difference in the post-processing of the U-net output. An additional equivariant convolution layer, followed by a group discretization operation, converts the output $\rho_{\text{irrep}}$ field into regular representations for discrete groups. The network $\phi_M$ encodes a zero-padded observation $\text{pad}(o_t) \in \mathbb{R}^{4 \times (H+p) \times (W+p)}$ into $\phi_M(\text{pad}(o_t)) \in \mathbb{R}^{|C_M| \times (H+p) \times (W+p)}$, an M-channel EOH map. Similarly, $\psi_N$ processes the image crop $c \in \mathbb{R}^{4 \times H_2 \times W_2}$ into an N-channel EOH map $\psi_N(c) \in \mathbb{R}^{|C_N| \times H_2 \times W_2}$. Using a subgroup alignment strategy, $\psi_N(c)$ is then transformed to a stack of N rotated M-channel EOH maps, denoted as $\mathcal{A}^N_M(\psi_N(c)) \in \mathbb{R}^{|C_N| \times |C_M| \times H_2 \times W_2}$. Cross-correlation between $\mathcal{A}^N_M(\psi_N(c))$ and $\phi_M(\text{pad}(o_t))$ is implemented as a convolution, with $\mathcal{A}^N_M(\psi_N(c))$ acting as the kernel. This operation produces the place distribution $p(a_{\text{place}}|o_t, a_{\text{pick}}) \in \mathbb{R}^{|C_N| \times H \times W}$.

*4) Training via behavior cloning.* The Histogram Transporter is trained using supervised learning on sequences of observation-action pairs $(o_t, \bar{a}_t)$ obtained from n = 1, 10 or 100 expert demonstrations per task. Each expert action $\bar{a}_t$ comprises a picking component $\bar{a}_{\text{pick}}$ and a placing component $\bar{a}_{\text{place}}$, both represented as binary one-hot pixel encodings. The picking and placing models are trained independently as action pose classifiers using cross-entropy loss. Model weights are updated over 10,000 iterations with the Adam optimizer, using a batch size of 1 and a fixed learning rate of $10^{-4}$. To enhance training robustness, random translation and rotation augmentations are applied to the sampled observation-action pair during each training step. Separate models are trained for each task, with performance evaluated every 2,000 iterations on 100 unseen configurations. Training typically completes within a few hours on a computing server equipped with an NVIDIA RTX 3090 Ti GPU and an AMD Ryzen Threadripper PRO 3995WX CPU.

## V. HAND-TOOL KITTING EXPERIMENTS

We systematically evaluate Histogram Transporter on hand-tool kitting tasks through both simulation and real-world experiments. The evaluation is divided into three core aspects: 1) performance comparison against baseline methods on HTKD (Subsection V-A), 2) ablation studies investigating factors affecting the method's performance (Subsection V-B), and 3) real-robot experiments validating the method's effectiveness in practical scenarios (Subsection V-C).

TABLE I
PERFORMANCE OF HISTOGRAM TRANSPORTER AND BASELINE METHODS ON HTKD (N = 180 ROTATIONS)

| Method | Memory Use (GBytes) ↓ | Inf. Time (Seconds) ↓ | kit-seen-toolset ↑ | | | kit-unseen-toolset ↑ | | |
| --- | --- | --- | --- | --- | --- | --- | --- | --- |
| | | | 1 | 10 | 100 | 1 | 10 | 100 |
| Histogram Transporter (ours) | **3.40** | 0.42 | **84** | **96** | 97 | **31.75** | **76.25** | **92** |
| SO(2)-Discretization | **3.40** | **0.38** | 79 | 90 | 96.50 | 29.25 | 72 | 87.25 |
| Equivariant Transporter | 17.41 | 0.58 | 83.25 | 91.75 | **97** | 28.25 | 75 | 88.75 |

### A. Comparison with Baselines on HTKD

*1) Experiment setup on HTKD.* We first evaluate the performance of Histogram Transporter in simulation using the HTKD benchmark [12]. As shown in Fig. 7, HTKD consists of 4 categories of hand tools: hammers, pliers, wrenches, and screwdrivers, each with 15 unique instances. The kitting process is simulated on PyBullet [37] using a Franka Emika gripper and three 640 × 480 RGB-D cameras positioned to overlook a $0.5 \times 1 \, m^2$ tabletop workspace from different viewpoints. While multiple cameras are used in simulation to improve visual coverage, our real-robot experiments confirm that the method works effectively with a single-camera setup.

We evaluate the proposed method on two hand-tool kitting tasks, designed to assess its generalizability under different object configurations:

- *kit-seen-toolset:* A single, fixed toolset-kit pair from the training pool is used across all training and testing runs, assessing generalizability to unseen object poses.
- *kit-unseen-toolset:* A toolset-kit pair is randomly sampled from the respective training and testing pools for each run, evaluating generalizability to unseen object instances and poses.

In both tasks, a set of four tools and their kit are randomly positioned in the workspace without collisions. The objective is to determine pick-and-place poses that allow the gripper to sequentially transport each tool into its designated cavity. To provide expert supervision, an oracle agent is developed for each task. This agent generates expert demonstrations, where each episode comprises exactly four observation-action pairs $(o_t, \bar{a}_t)$, corresponding to the sequential kitting of all four tools.

Performance is evaluated based on task success rate and computational efficiency. A kitting action is considered successful only if the tool fully fits into its corresponding cavity, and place accuracy is quantified using the Average Distance Metric for Symmetric Objects (ADD-S) [38], a standard metric in pose estimation. Each test episode involves kitting four tools, and the overall success rate is computed on a scale from 0 (failure) to 100 (success), with 25 points awarded per successful kitting action. Presuming one action per tool, each test episode is limited to four kitting steps. The reported results are the highest validation results achieved during training, averaged over 100 unseen test episodes per task. In addition to success rate, we measure computational efficiency in terms of memory footprint and inference time during evaluation. For more details on the task setup, please refer to [12].

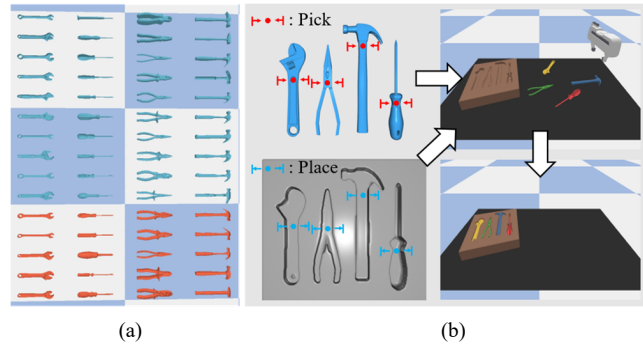

Fig. 7. Hand-Tool Kitting Dataset. (a) Tool models. HTKD includes 4 tool categories: wrenches, screwdrivers, pliers, and hammers, with 15 instances per category (blue for training, red for testing). (b) Expert demonstration: A simulation example involving a pair of tools and their kit. The gripper executes the annotated pick-and-place actions to complete the kitting task.

We compare Histogram Transporter against two state-of-the-art baselines on HTKD: Equivariant Transporter [9] and SO(2)-Discretization [12]. The *Equivariant Transporter* employs a similar architecture but uses $C_N$ regular representations in its intermediate layers and pools invariant feature embeddings (trivial representations) for cross-correlation in the place model $f_{place}$. The *SO(2)-Discretization* baseline shares the same model architectures and irreducible representations as Histogram Transporter but pools irreducible representations into invariant trivial representations for cross-correlation in $f_{place}$. To ensure a fair comparison, all methods follow the same training and evaluation procedure, as outlined in Subsection IV-D. The default rotation number N is empirically set to 180 for all methods to ensure sufficient orientation precision, unless otherwise specified. Similarly, the subgroup size $|C_M|$ for subgroup alignment is set to M = 12 by default.

*2) Results.* Table I presents the success rates and computational efficiency of the proposed Histogram Transporter compared to baseline approaches across two kitting tasks, trained with varying numbers of demonstrations. Histogram Transporter achieves the highest success rates in all scenarios. For the kit-seen-toolset task, both Histogram Transporter and Equivariant Transporter achieve a 97% success rate after training on 100 demonstrations. However, Histogram Transporter demonstrates superior sample efficiency, achieving a 96% success rate with just 10 demonstrations — a notable improvement over the baselines. For the more challenging kit-unseen-toolset task, Histogram Transporter consistently outperforms all baselines at every demonstration level, validating its strong generalization to novel tool instances. The improvement over SO(2)-Discretization stems from incorporating EOH encodings in $f_{\text{place}}$. These results highlight the critical role of EOH in driving Histogram Transporter's superior accuracy and generalizability over Equivariant Transporter.

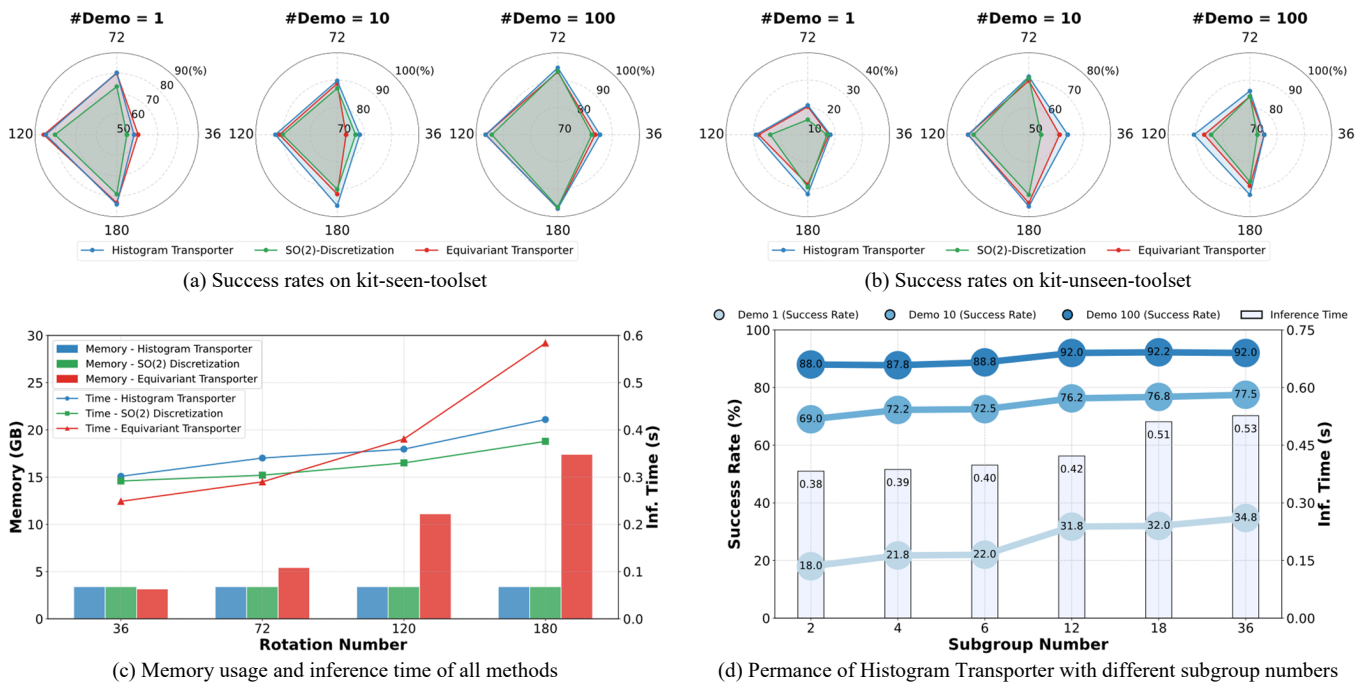

Fig. 8. Ablation study on rotation and subgroup alignment numbers. (a) and (b) present radar charts comparing the success rates of Histogram Transporter against SO(2)-Discretization and Equivariant Transporter on HTKD under different rotation numbers and demonstration counts for seen and unseen toolsets, respectively. In each radar chart, the four directions represent different rotation numbers, while the colored lines indicate different methods. (c) contrasts memory usage (bars) and inference time per test run (lines) during evaluation. (d) examines the impact of different subgroup alignment settings on the kit-unseen-toolset task, where success rates are shown as lines, with darker shades indicating higher demonstration numbers, and inference times depicted as bars.

Beyond its high success rates, Histogram Transporter is significantly more computationally efficient than Equivariant Transporter, benefiting from the Fourier-based EOH generation method. Specifically, it requires 5.1× less memory and infers 1.4× faster. While its inference time is slightly longer than that of SO(2)-Discretization due to the added computations from EOH encodings in $f_{\text{place}}$, Histogram Transporter strikes an effective balance, offering substantially higher success rates while maintaining competitive memory usage and inference speed.

*B. Ablation study*

We also conduct ablation studies to examine key factors influencing Histogram Transporter's performance, including: 1) the number of orientation bins in the action space, 2) the subgroup size used in the subgroup alignment strategy, and 3) the individual contributions of the pick and place modules.

*1) Orientation numbers in action space.* We investigate the impact of angular resolution in the action space on kitting performance by experimenting with orientation numbers N = 36, 72, 120 and 180. Figs. 8(a) and 8(b) illustrate how varying angular resolutions affect success rates on HTKD across different methods and demonstration counts. Higher angular resolutions generally lead to improved success rates, with N = 180 yielding the highest scores across all methods. These results confirm the orientation sensitivity inherent in the HTKD tasks, highlighting the importance of high-precision actions. Furthermore, our method outperforms the baseline methods in nearly all configurations, demonstrating its robustness in kitting applications.

Fig. 8(c) illustrates the memory usage and inference time for all methods across different rotation numbers. While all methods require similarly low memory (~4GB) at N = 36, their behaviors diverge as the rotation space increases. Histogram Transporter and SO(2)-Discretization maintain consistent memory usage regardless of the rotation number. In contrast, the memory footprint of Equivariant Transporter scales linearly with the size of rotation space, leading to a substantial disparity at higher resolutions, such as 20 GB vs. 4 GB at N = 180. This rapid growth imposes constraints on GPU memory capacity, therefore limiting the scalability of Equivariant Transporter in larger rotation space.

These divergences in memory usage primarily arise from differences in their pick angle networks $f_\theta$. Specifically, the Equivariant Transporter uses $C_N/C_2$ regular representations in its intermediate levels, requiring additional parameters to explicitly encode the orientation information of larger rotation groups. In contrast, Histogram Transporter and SO(2)-Discretization rely on a fixed set of irreducible representations to approximate the entire rotation space, rendering their model parameters independent of rotation numbers. This design difference also impacts inference time, with the Equivariant Transporter being much slower at large rotation spaces. Moreover, the increase in inference time across all methods as rotation spaces expand is due to the growing stack size of rotated templates during the cross-correlation step in $f_{\text{place}}$.

TABLE II
PERFORMANCE OF DIFFERENT TRANSPORTER CONFIGURATIONS ON HTKD

| Method | EOH-based Picking | EOH-based Placing | Memory Use (GBytes) ↓ | kit-seen-toolkit ↑ | | | kit-unseen-toolkit ↑ | | |
|---|---|---|---|---|---|---|---|---|---|
| | | | | 1 | 10 | 100 | 1 | 10 | 100 |
| Equivariant Transporter | × | × | 17.41 | 83.25 | 91.75 | 97 | 28.25 | 75 | 88.75 |
| Hybrid 1 | ✓ | × | **3.22** | 73.75 | 89.25 | 93 | 21.5 | 69 | 89.25 |
| Hybrid 2 | × | ✓ | 17.68 | **84.75** | 92.5 | **97.75** | **34.5** | **79** | 91.25 |
| Histogram Transporter (Ours) | ✓ | ✓ | 3.40 | 84 | **96** | 97 | 31.75 | 76.25 | **92** |

*2) Subgroup Size in Subgroup Alignment Strategy.* We examine how the subgroup size in the subgroup alignment strategy influences task success rates and inference speed by creating six place model variations with subgroup sizes M = 2, 4, 6, 12, 18 and 36. Fig. 8(d) reports the highest validation results for all model variations across different demonstration numbers. Success rates initially increase with larger subgroup sizes and plateau after M = 12. The most significant improvement occurs when transitioning from $C_6$ to $C_{12}$, as smaller subgroups ($C_M \leq C_6$) suffer from aliasing effects that violate the sampling rule. Beyond $C_{12}$, further increases in subgroup size yield minimal improvement in success rates but substantially increase processing time. As a result, we select $C_M = C_{12}$ as the default subgroup size to optimize the trade-off between success rates and computational efficiency.

*3) Module Swap Evaluation.* To study the individual contributions of the pick and place modules to the overall kitting performance, we perform a module swap experiment. Specifically, components are interchanged between our method and the Equivariant Transporter, resulting in four configurations: 1) Histogram Transporter, 2) Equivariant Transporter, 3) Hybrid I (Hist. pick and Equi. place) and 4) Hybrid II (Equi. pick and Hist. place).

As shown in Table II, our proposed place module consistently improves success rates regardless of the pick module adopted, underscoring the effectiveness of our EOH descriptors. Combining the Equivariant Transporter pick module and our place module, Hybrid II delivers the highest success rates in 4 out of the 6 test configurations, albeit at the cost of reduced computational efficiency. Conversely, the Histogram Transporter achieves a superior balance by integrating an efficient pick model with an accurate place model, delivering both high accuracy and scalability for orientation-sensitive kitting tasks.

*C. Real-World Kitting Experiments*

We further evaluate the performance of the Histogram Transporter on a real-world kitting setup. For this experiment, the model is trained from scratch using demonstration data collected from the physical robot.

*1) Experiment Setup.* Our real-robot setup consists of a Universal Robot UR10e equipped with a Robotiq 2F-85 gripper and an industrial Zivid One+ M camera, as shown in Fig. 9. The overhead-mounted camera captures both RGB and depth images of a tabletop surface, spanning a 0.4 × 0.6 m² area. The captured images are used to generate a heightmap with a resolution of 160 × 240 pixels, corresponding to a spatial resolution of 2.5 mm per pixel. Based on the action poses predicted by the Histogram Transporter, the robot motions are executed using MoveIt!.

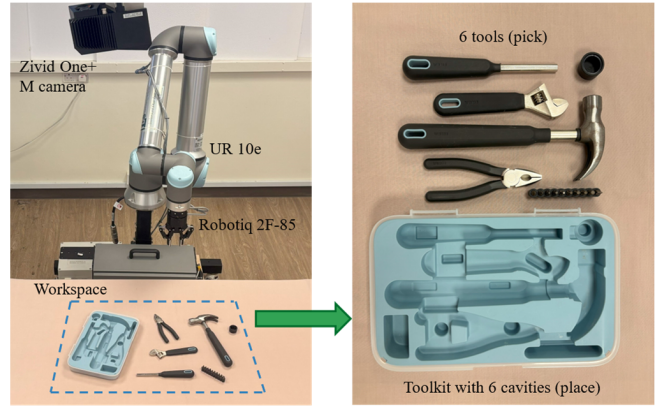

Fig. 9. Real-robot setup (left) and IKEA toolset for kitting (right).

We adopt a real-robot implementation of the simulated kit-seen-toolset task from HTKD. The task is conducted with an off-the-shelf IKEA toolbox containing six tool categories: a screwdriver bit set, a hammer cap and the 4 basic tool types from HTKD, as illustrated in Fig. 9. The two additional tools introduces further challenges due to their small sizes and distinct shapes.

In this experiment, we collect a dataset of 10 human demonstrations using Cartesian servoing via the robot's control panel. Without any sim-to-real transfer or pretraining, a single Histogram Transporter model is trained on the collected data for 10,000 SGD steps, using the same optimizer, batch size and learning rate as in the simulation experiments. The trained model is subsequently evaluated on 20 unseen test configurations, where the tools and the kit are randomly positioned over the table surface.

*2) Results.* Table III summarizes the kitting results for each tool instance and the overall set completion rate. The Histogram Transporter exhibits outstanding performance, successfully completing 19 out of 20 kitting sets during evaluation. Fig. 10 illustrates a successful kitting sequence, with more qualitative results available in the supplementary videos. Among the six tools, five achieve a 100% success rate. Interestingly, the overall success rate slightly exceeds that of the simulated task, likely due to vibrations during tool placement, which assist in the tools' entries into their cavities.

TABLE III
SUCCESS RATES OF INDIVIDUAL TOOLS IN THE REAL-ROBOT KITTING EXPERIMENT

| Tool<br>S. R. (%) | Hammer<br>100 (20/20) | Screwdriver<br>100 (20/20) | Wrench<br>100 (20/20) | Pliers<br>100 (20/20) | Hammer Cap<br>100 (20/20) | Scrdri. Bits<br>95 (19/20) | Entire Set<br>95 (19/20) |
|---|---|---|---|---|---|---|---|

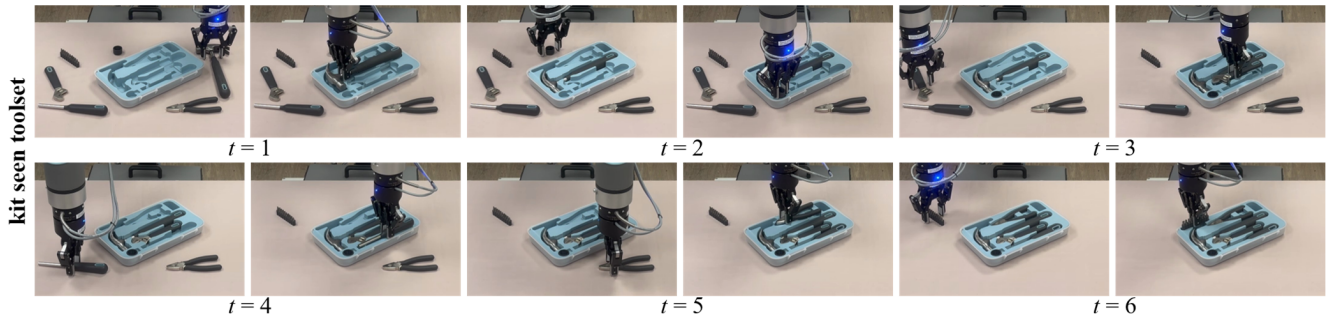

Fig. 10. Snapshots of the real-world hand-tool kitting experiment. Histogram Transporter generates a sequence of 6 accurate pick-and-pace poses to kit all randomly positioned hand tools into their corresponding cavities.

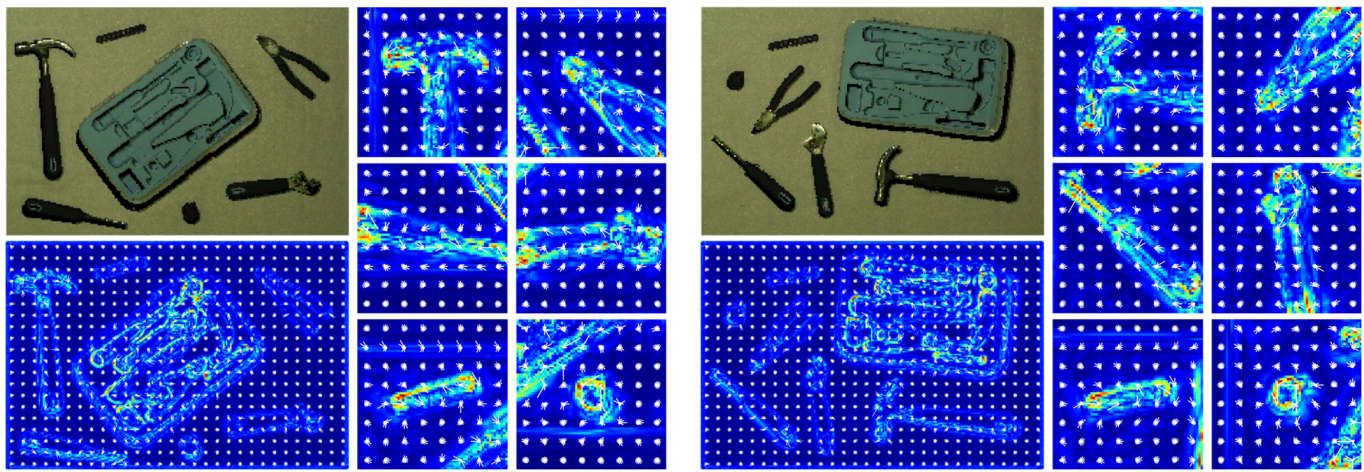

Fig. 11. Visualization of the learned EOH encodings from $f_{\text{place}}$ for two test episodes. The workspace encodings (bottom left) are captured at the start of each test episode, while the tool encodings are captured sequentially at their respective kitting rounds. The EOHs are visualized as multi-directional arrows at a stride of 8, with the pixel color indicating the maximal magnitude of the EOH at the corresponding pixel.

The only failure occurs with the screwdriver bit set, where a minor placement error causes the bit set to get stuck at the cavity entrance due to friction. This issue arises from the bit set's thin, uneven structures, which provide few points in the heightmap regarding its color and shape. Moreover, a notable case is observed with the wrench: despite the demonstrated actions focusing on the middle part of the wrench, the robot successfully completes placement twice by grasping the end of the wrench handle instead. This novel kitting pose highlights the robustness of the place model, which effectively generalizes to unseen pick locations rather than overfitting to the demonstrated actions.

*3) Feature Visualization.* To gain deeper insight into the learned EOHs, we visualize the histogram values for two test episodes in Fig. 11. The EOHs are shown as multi-directional arrows with a stride of 8, and pixel colors represent the maximal orientation value of each EOH. These visualization reveal the following properties: 1) objectness detection: foreground objects exhibit significantly higher major orientation values than the background table surface, indicating high-variance EOHs for objects and low-variance EOHs for the table; 2) object differentiation: different tools display unique descriptor patterns around their shapes; 3) spatial correspondence: corresponding points on the same object maintain similar descriptor values for the major orientations, irrespective of the object's orientation; and 4) cavity detection: the cavity contours show prominent major orientations, providing crucial information for placing targets.

## VI. ADDITIONAL PICK-AND-PLACE EVALUATION

This section explores the adaptability of the Histogram Transporter to diverse pick-and-place tasks beyond kitting. Subsection VI-A benchmarks our method against baselines on five tasks from the Raven-10 platform. Subsection VI-B extends this evaluation to real-world implementations of two of these tasks using a physical robot.

### A. Simulation Experiments.

*1) Experiment Setup.* We evaluate our method on five tasks from the Raven-10 benchmark, modified to work with a

TABLE IV
SUCCESS RATES OF HISTOGRAM TRANSPORTER AND BASELINE METHODS ON RAVEN-10 (N = 36 ROTATIONS)

| Method | block-insertion | | | place-red-in-green | | | palletizing-boxes | | | align-box-corner | | | stack-block-pyramid | | |
|---|---|---|---|---|---|---|---|---|---|---|---|---|---|---|---|
| | 1 | 10 | 100 | 1 | 10 | 100 | 1 | 10 | 100 | 1 | 10 | 100 | 1 | 10 | 100 |
| Histogram Transporter | **100** | **100** | **100** | 92.8 | **100** | **100** | **98.4** | 99.9 | **100** | **66.0** | 99.0 | **100** | 60.2 | **92.8** | **99.2** |
| SO(2)-Discretization | **100** | **100** | **100** | **96.0** | **100** | **100** | 97.8 | 99.9 | **100** | 58.0 | **100** | **100** | 51.8 | 84.5 | 97.5 |
| Equivariant Transporter | **100** | **100** | **100** | 95.6 | **100** | **100** | 96.1 | **100** | **100** | 64.0 | 99.0 | **100** | **62.1** | 85.6 | 98.3 |
| Transporter Network | 98.0 | **100** | **100** | 82.3 | 94.8 | **100** | 84.2 | 99.6 | **100** | 45.0 | 85.0 | 99.0 | 16.6 | 63.3 | 75.0 |

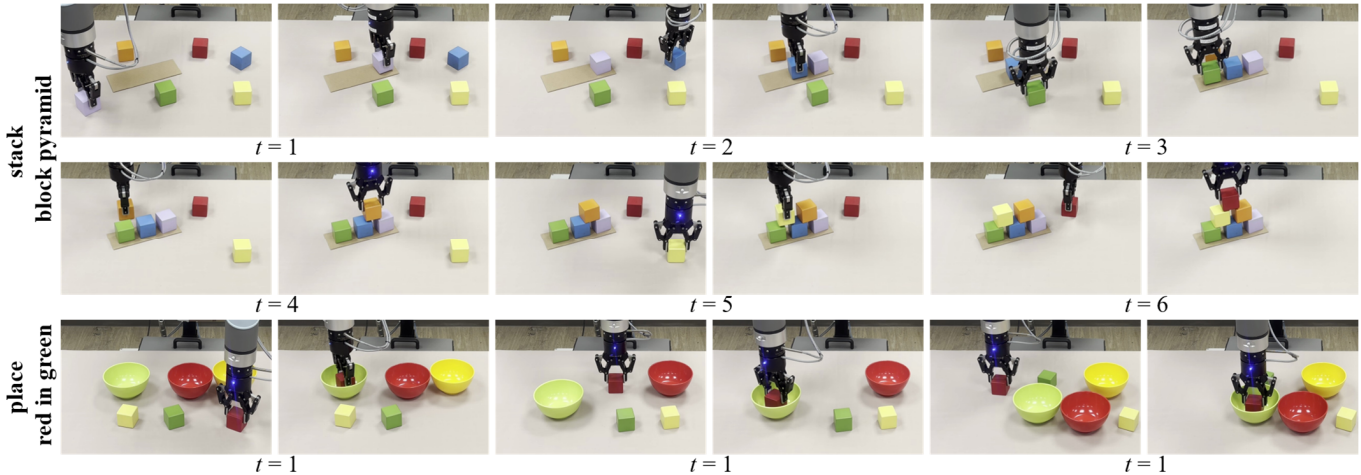

Fig. 12. Snapshots of real-robot experiments for the stack-block-pyramid and place-red-in-green tasks. The first two rows illustrate the sequential pick-and-place actions for stacking six blocks. The third row presents three test episodes of the place-red-in-green task, each consisting of a single pick-and-place step.

parallel-jaw gripper, following the setup of the Equivariant Transporter. The tasks include:

- *block-insertion*: picking up an L-shaped block and placing it to fill a corresponding hole in an L-shaped fixture.
- *align-box-corner*: picking up a box of random size and placing it to align one of its corners to a green L-shape marker on the table.
- *place-red-in-green*: picking up all red blocks on the table and placing them into green bowls.
- *stack-block-pyramid*: sequentially picking up six blocks of different colors and stacking them into a pyramid of 3-2-1.
- *palletizing-boxes*: picking up eighteen boxes of identical size and stacking them onto a pallet.

Task performance is evaluated using two metrics. The palletizing-boxes task measures the ratio of objects placed in the target zone. For the other tasks, performance is evaluated based on the translation and rotation errors of the objects relative to their target pose. Further task configuration details can be found in [9]. Since these tasks require less orientation precision compared to kitting, we adopt a small rotation number (N = 36), aligning with prior works [8], [9]. Apart from this adjustment, the training and evaluation procedures remain consistent with those used in the HTKD experiments.

In addition to the Equivariant Transporter and SO(2)-Discretization baselines from the kitting experiments (see Subsection V-A), we include another baseline introduced in [8]. This adaptation integrates 43-layer ResNets built with conventional convolution layers and is tailored for a parallel-jaw gripper. Specifically, the input observation $o_t$ is lifted to $|C_N|$ rotations, forming a stack of oriented images as input for the pick network $f_{\text{pick}}$. Performance results for the Equivariant Transporter and the adapted Transporter Network on Raven-10 are taken from [9].

*2) Results.* Table IV outlines the success rates of all methods across five tasks with different numbers of demonstrations for training. Our method closely matches the performance of state-of-the-art approaches across all tasks. Specifically, it attains a success rate of ≥ 99% on 4 out of 5 tasks with just 10 demonstrations and on all tasks with 100 demonstrations. These results highlight the effectiveness of our method in handling a variety of pick-and-place tasks with limited demonstration data, making it ideal for applications requiring high sample efficiency and rapid adaptability.

*B. Real-World Experiments*

We further evaluate our method on the *place-red-in-green* and *stack-block-pyramid* tasks using the same real-robot platform as the kitting tasks, illustrated in Fig. 12. For each task, we collect a dataset of 10 demonstrations and train a single-task policy from scratch for 10,000 iterations. The trained models are then tested on 20 unseen configurations, with the results summarized in Table V. Our method achieves a 100% success rate in the place-red-in-green task and a 90% success rate in the stack-block-pyramid task. The two observed failure modes in the latter include a missed placement of the purple block onto the pallet and an excessive gap between stacked blocks for subsequent placement. These real-world results demonstrate that the sample-efficiency observed in simulation translates effectively to physical implementations, showcasing the feasibility of training high-performance policies from scratch in real-world scenarios.

TABLE V
SUCCESS RATES OF TWO RAVEN-10 TASKS ON THE REAL-ROBOT PLATFORM

| Task | Success Rate (%) |
|---|---|
| place-red-in-green | 100 (20/20) |
| stack-block-pyramid | 90 (18/20) |

## VII. Conclusion

In this work, we present Histogram Transporter, an imitation learning framework designed to acquire high-precision kitting policies from a limited number of demonstrations. Central to our method is the dual application of EOHs, which empower our kitting framework with an accurate and computationally efficient architecture for orientation-sensitive kitting tasks while ensuring data efficiency and generalizability. Comprehensive experiments in simulation and real-world settings validate the accuracy, scalability, and adaptability of the Histogram Transporter.

Beyond robotic kitting, our work has broader implications for robotic manipulation and computer vision. First, experiments on the Raven-10 tasks demonstrate that Histogram Transporter can adapt to a wide range of manipulation tasks with varying complexities. Second, the distinct roles of EOHs in the pick and place modules align with two classical vision challenges: orientation estimation and feature description. This connection opens up opportunities for additional applications, such as oriented object detection and image matching. Finally, by leveraging a truncated Fourier representation to approximate continuous signals over the entire rotation space, our method provides an efficient and effective alternative to explicit modeling of discrete rotation signals.

While our method shows promising results, several limitations suggest directions for future exploration. First, the current approach is limited to planar kitting tasks; extending it to more complex 6-DoF kitting scenarios is a crucial future direction. Second, although our evaluation confirms generalizability to unseen object poses and instances within known categories, developing a generalized kitting policy capable of handling entirely novel objects remains an exciting challenge. Lastly, the HTKD is used solely for evaluation purposes. Future work could investigate sim-to-real transfer to enable seamless deployment of models trained in simulation to real-world applications.